\documentclass{article}

\usepackage[final,nonatbib]{neurips_2021}

\usepackage[utf8]{inputenc} 
\usepackage[T1]{fontenc}    
\usepackage{hyperref}       
\usepackage{url}            
\usepackage{booktabs}       
\usepackage{amsfonts}       
\usepackage{nicefrac}       
\usepackage{microtype}      

\usepackage{multirow}
\usepackage[table,xcdraw]{xcolor}
\usepackage{graphicx}
\usepackage{booktabs}
\usepackage{subfig}

\title{Multi-Domain Balanced Sampling Improves Out-of-Distribution Generalization of Chest X-ray Pathology Prediction Models}

\author{Enoch Tetteh \\
  Mila, Quebec AI Institute \\
  AMMI, AIMS Rwanda \\
  \texttt{etetteh@aimsammi.org} \\
    \And
  Joseph Viviano \\
  Mila, Quebec AI Institute \\
  University of  \\
    \And
  Yoshua Bengio \\
  Mila, Quebec AI Institute \\
  University of Montreal \\
    \And 
  David Krueger \\
  Mila, Quebec AI Institute \\
  University of Montreal \\
  University of Cambridge \\
    \And
  Joseph Paul Cohen \\
  Mila, Quebec AI Institute \\
  AIMI, Stanford University \\
}

\begin{document}

\maketitle

\begin{abstract}
  
  Learning models that generalize under different distribution shifts in medical imaging has been a long-standing research challenge. 
  There have been several proposals for efficient and robust visual representation learning among vision research practitioners, especially in the sensitive and critical biomedical domain.
  In this paper, we propose an idea for out-of-distribution generalization of chest X-ray pathologies that uses a simple balanced batch sampling technique.  
  We observed that balanced sampling between the multiple training datasets improves the performance over baseline models trained without balancing.
  Code for this work is available on Github.\footnote{\url{https://github.com/etetteh/OoD_Gen-Chest_Xray}}
\end{abstract}

\section{Introduction}
Pathology detection or classification in medical imaging using deep learning \cite{Hao2016DeepL}, \cite{Goodfellow2015DeepL} continues to be an open research challenge. Although the field of computer vision has progressed significantly owing to advances in model architecture, optimization techniques, and data augmentation, learning relevant feature representations from visual data still remains a challenge \cite{Cohen2020OnTL} \cite{MorenoTorres2012AUV}. 
The field of medical imaging using deep learning suffers from the same issues, since they rely mostly on the same deep learning approaches inspired by studies on general purpose datasets, like ImageNet \cite{Russakovsky2015ImageNetLS}. The issue becomes more challenging when test data is subject to distribution shifts \cite{QuioneroCandela2009DatasetSI}. 
When confronted with chest X-rays from different datasets, a straightforward approach, followed by \cite{Cohen2020OnTL}, is to 
merge these datasets and 
form mini-batches by sampling uniformly from the merged dataset.


The question is: can we find a more robust and efficient way of learning representations from medical images by accounting for which dataset each example came from?

In this work, we examine how balanced batch sampling from each training dataset can improve a model's generalization on out-of-distribution (OoD) chest X-ray datasets. We compare this to \cite{Cohen2020OnTL} and a baseline model trained without balanced batching.

Our work shows that, by training vision algorithms on chest X-rays using balanced mini-batches, we may achieve performance gains during inference on out-of-distribution chest X-ray datasets.

\section{Experiment}
We aim to classify 4 chest X-ray pathologies, namely Cardiomegaly, Consolidation, Edema, and Effusion.
We compare our approach to two baselines that follow SOTA medical image classification approaches. We describe the baselines 
in ~\ref{section:training}.

This choice of pathologies is strictly because all the datasets in this experiment include labels for all of these pathologies, and not for any other particular reason.
The training, validation and test datasets all come from different distributions. This is to ensure that test and validation datasets are out-of-distribution with respect to the training data. For example, when training on ChestX-ray8 and CheXpert datasets, we validate on MIMIC-CXR dataset and test on PadChest.  

\paragraph{Datasets}
We consider four publicly available chest X-ray image datasets for this experiment: ChestX-ray8 \cite{Wang2017ChestXRay8HC},
CheXpert \cite{Irvin2019CheXpertAL}, MIMIC-CXR-JPG \cite{Johnson2019MIMICCXRAL}, and PadChest \cite{Bustos2020PadChestAL}. Table \ref{tab:datasets} displays the details of our datasets.

\begin{table}[!h]
\caption{Details of datasets in this work. \newline}
\label{tab:datasets}
\resizebox{1\textwidth}{!}{
\centering
\begin{tabular}{lcccc}
\toprule
\textbf{Datasets}   & \textbf{\# Samples Total} & \textbf{\# Samples Used} & \textbf{\# Pathologies} & \textbf{Geo.   Region} \\ 
& (unique patients) \\
\midrule
ChestX-ray8 dataset (NIH) & 67,310        &      27,520        & 14                       & Northeast USA     \\ 
MIMIC-CXR Dataset (MIMIC)  & 96,155        &      27,520       & 13                       & Northeast USA     \\ 
PadChest Dataset (PC)   & 91,658        &      27,520       & 27                       & Spain             \\ 
CheXpert Dataset (CHEX)   & 29,420        &      27,520       & 13                       & Western USA       \\
\bottomrule
\end{tabular}
}
\end{table}

We use a subset of 27,520 images from each dataset for training, 11,008 for validation, and 13,760 for inference. We end up with training samples of 55,040 images, since we use two datasets as the train set. The training data is sampled sequentially from the original dataset. 
On the other hand, the entire validation and test datasets are sampled from the data loaders as batches. 
We use the TorchXRayVision \cite{Cohen2020OnTL, Cohen2020xrv} library, which is specialized in handling chest X-ray images, to load the raw data.

\subsection{Training}
\label{section:training}
In order to increase the diversity of the training dataset to improve the robustness of the trained model \cite{Wong2016UnderstandingDA}, we perform the following basic augmentation techniques: we resize  all images to 112x112 resolution (this is done to speed up the experiments), rotate the images by up to 45 degrees, translate by up to [-0.15, 0.15], and scale by [0.85, 1.15].

We perform a leave-a-dataset-out cross validation, and perform training using a DenseNet-121 architecture from the torchvision library. The model is fine-tuned \cite{Pan2010ASO, Yosinski2014HowTA} using ImageNet weights, and without any modification to its architecture except the input channel is changed to 1 (for gray-scale images). 

We use an Adam \cite{Kingma2015AdamAM} optimizer with a fixed learning rate of 1e-03, weight decay of 1e-05, and amsgrad set to true. We run all our experiments for 200 epochs with a batch size of 64, using a binary cross-entropy with logits as our loss function.
We perform early stopping \cite{Prechelt2012EarlyS}, and use the final best validation model state, for our inference.

We run the experiments with three different seeds and report the average. Unless otherwise stated, all results are on the test datasets.

\subsection{Models compared}

We compare with 2 different baselines, one taken from previous work and one which we train following \cite{Cohen2020OnTL}, a state-of-the-art method, on our specific datasets:

\noindent\textbf{Baseline XRV} is a DenseNet-121 model, from the TorchXRayVision library \cite{Cohen2020xrv} that is trained on 8 chest X-ray datasets (including all the datasets we use in our experiments), and tasked to classify 18 chest X-ray pathologies. We only performed inference using this model for comparison with our models.

\noindent\textbf{Random Batch Sampling} follows previous approaches of studies performed on chest X-ray pathology classification using different data distributions, which involves merging multiple datasets into a larger one for training. We merge two datasets for training, and the remaining two are use for validation and inference respectively.

\noindent\textbf{Balanced Batch Sampling}
We create two training environments, one for each of our training datasets. At each training step, we sample data from each of the environments, compute the individual losses and back-propagate using the sum of the losses from the environments.

\section{Results and Discussion}
\label{section:Results and Discussion}
In this section, we present and discuss the findings of our work. Table \ref{tab:results} shows the results of our experiments.

Our results suggest that out-of-distribution generalization performance may be improved by using a balanced batching technique - sampling data from each environment/dataset equally and computing the sum of the losses.

\begin{table}[!h]
\caption{Comparison of balanced batch sampling to not balanced and an existing pre-trained model.\\}
\label{tab:results}
\resizebox{1\textwidth}{!}{
\centering
\begin{tabular}{
>{\columncolor[HTML]{FFFFFF}}c |
>{\columncolor[HTML]{FFFFFF}}l |
>{\columncolor[HTML]{FFFFFF}}c 
>{\columncolor[HTML]{FFFFFF}}c 
>{\columncolor[HTML]{FFFFFF}}c 
>{\columncolor[HTML]{FFFFFF}}c 
>{\columncolor[HTML]{FFFFFF}}c 
>{\columncolor[HTML]{FFFFFF}}c |
>{\columncolor[HTML]{FFFFFF}}c }
\multicolumn{1}{l|}{\cellcolor[HTML]{FFFFFF}} & Seed & \textbf{0} & \textbf{42} & \textbf{99} & \textbf{1} & \textbf{2} & \textbf{3} & \multicolumn{1}{l}{\cellcolor[HTML]{FFFFFF}} \\ \cline{2-8}
\textbf{Model} & \begin{tabular}[c]{@{}l@{}}TRAIN\\ VALID\\ TEST\end{tabular} & \textbf{\begin{tabular}[c]{@{}c@{}}NIH\_CHEX\\ MIMIC\\ PC\end{tabular}} & \textbf{\begin{tabular}[c]{@{}c@{}}NIH\_PC\\ MIMIC\\ CHEX\end{tabular}} & \textbf{\begin{tabular}[c]{@{}c@{}}CHEX\_MIMIC\\ PC\\ NIH\end{tabular}} & \textbf{\begin{tabular}[c]{@{}c@{}}NIH\_MIMIC\\ CHEX\\ PC\end{tabular}} & \textbf{\begin{tabular}[c]{@{}c@{}}CHEX\_PC\\ NIH\\ MIMIC\end{tabular}} & \textbf{\begin{tabular}[c]{@{}c@{}}MIMIC\_PC\\ CHEX\\ NIH\end{tabular}} & \textbf{MEAN} \\ \toprule
\cellcolor[HTML]{FFFFFF} & \textit{\textbf{Best Valid AUC}} & - & - & - & - & - & - & - \\
\cellcolor[HTML]{FFFFFF} & \textit{\textbf{Avg Test AUC}} & \textbf{0.91} & \textbf{0.92} & \textbf{0.82} & \textbf{0.91} & \textbf{0.92} & \textbf{0.82} & \textbf{0.88} $\pm$ \textbf{0.05} \\
\cellcolor[HTML]{FFFFFF} & \textit{\textbf{Cardiomegaly}} & 0.92 & 0.90 & 0.91 & 0.92 & 0.90 & 0.91 & 0.91 $\pm$ 0.01 \\
\cellcolor[HTML]{FFFFFF} & \textit{\textbf{Effusion}} & 0.94 & 0.95 & 0.86 & 0.94 & 0.95 & 0.86 & 0.92 $\pm$ 0.04 \\
\cellcolor[HTML]{FFFFFF} & \textit{\textbf{Edema}} & 0.95 & 0.92 & 0.76 & 0.95 & 0.92 & 0.76 & 0.87 $\pm$ 0.09 \\
\multirow{-6}{*}{\cellcolor[HTML]{FFFFFF}\textbf{\begin{tabular}[c]{@{}c@{}}*Baseline XRV \\ (no training, \\ using pre-trained\\ `all' model from \cite{Cohen2020xrv})\end{tabular}}} & \textit{\textbf{Consolidation}} & 0.85 & 0.90 & 0.75 & 0.85 & 0.90 & 0.75 & 0.83 $\pm$ 0.07 \\ \midrule
\cellcolor[HTML]{FFFFFF} & \textit{\textbf{Best Valid AUC}} & 0.85 & 0.84 & 0.89 & 0.81 & 0.79 & 0.88 & 0.84 $\pm$ 0.04 \\
\cellcolor[HTML]{FFFFFF} & \textit{\textbf{Avg Test AUC}} & \textbf{0.89} & \textbf{0.80} & \textbf{0.74} & \textbf{0.87} & \textbf{0.89} & \textbf{0.76} & \textbf{0.82} $\pm$ \textbf{0.07} \\
\cellcolor[HTML]{FFFFFF} & \textit{\textbf{Cardiomegaly}} & 0.92 & 0.81 & 0.79 & 0.92 & 0.88 & 0.78 & 0.85 $\pm$ 0.07 \\
\cellcolor[HTML]{FFFFFF} & \textit{\textbf{Effusion}} & 0.90 & 0.85 & 0.80 & 0.89 & 0.93 & 0.81 & 0.86 $\pm$ 0.05 \\
\cellcolor[HTML]{FFFFFF} & \textit{\textbf{Edema}} & 0.94 & 0.78 & 0.70 & 0.91 & 0.90 & 0.74 & 0.83 $\pm$ 0.10 \\
\multirow{-6}{*}{\cellcolor[HTML]{FFFFFF}\textbf{\begin{tabular}[c]{@{}c@{}}Random Batching\\ (finetuned from \\ ImageNet model)\end{tabular}}} & \textit{\textbf{Consolidation}} & 0.79 & 0.77 & 0.66 & 0.77 & 0.86 & 0.69 & 0.76 $\pm$ 0.07 \\ \midrule
\cellcolor[HTML]{FFFFFF} & \textit{\textbf{Best Valid AUC}} & 0.91 & 0.89 & 0.91 & 0.89 & 0.81 & 0.89 & 0.88 $\pm$ 0.04 \\
\cellcolor[HTML]{FFFFFF} & \textit{\textbf{Avg Test AUC}} & \textbf{0.90} & \textbf{0.86} & \textbf{0.79} & \textbf{0.91} & \textbf{0.89} & \textbf{0.80} & \textbf{0.86} $\pm$ \textbf{0.05} \\
\cellcolor[HTML]{FFFFFF} & \textit{\textbf{Cardiomegaly}} & 0.90 & 0.87 & 0.83 & 0.92 & 0.88 & 0.87 & 0.88 $\pm$ 0.03 \\
\cellcolor[HTML]{FFFFFF} & \textit{\textbf{Effusion}} & 0.92 & 0.92 & 0.83 & 0.93 & 0.93 & 0.84 & 0.89 $\pm$ 0.05 \\
\cellcolor[HTML]{FFFFFF} & \textit{\textbf{Edema}} & 0.95 & 0.85 & 0.79 & 0.95 & 0.90 & 0.79 & 0.87 $\pm$ 0.07 \\
\multirow{-6}{*}{\cellcolor[HTML]{FFFFFF}\textbf{\begin{tabular}[c]{@{}c@{}}Balanced Batching\\ (finetuned from\\  ImageNet model)\end{tabular}}} & \textit{\textbf{Consolidation}} & 0.84 & 0.82 & 0.72 & 0.85 & 0.86 & 0.71 & 0.80 $\pm$ 0.07
\end{tabular}}
{\textsuperscript{*}}\footnotesize Trained using a superset of data compared to models trained in our work. Best Valid AUC is not reported, because the model was used off-the-shelf for inference only.
\end{table}

From Table \ref{tab:results}, we can observe performance gains from the balanced batch sampling model over the random batch sampling model. By using balanced batching, we are able to outperform  the random sampling approach in all six experimental settings, in this work. 

Randomly sampling mini-batches from a merged dataset may result in data bias, because the sampled data may come from only one of the multiple distributions available or there may be fewer samples from some distributions than the other. This may result in a model biased towards a certain distribution. On the other hand, the balanced batch sampling uses a stratified sampling approach, which ensures the algorithm sees data from each distribution at every iteration during training. The model in this case is less/not biased towards any of the distribution.
The training datasets themselves were balanced for both balanced and random batch sampling. So it appears the overall balancing is not as impactful as the balancing of the mini-batches passed to the algorithm.
The challenge of sample imbalance for each task is taken care of by computing a weighted loss.

Also, although the Baseline XRV model is trained on a much larger data, and overlaps the test data, the average AUC of our model trained with balanced batching is as almost good as the XRV model.   

\section*{Potential negative societal impact}
\label{section:Potential negative societal impact}
This research uses only previously public data, so there are no privacy concerns. We do not foresee any negative societal impact as a result of the research described in this work.

\section*{Acknowledgments}
\label{section:acknowledgments}
This research is based on work partially supported by the CIFAR AI and COVID-19 Catalyst Grants. We would like to thank Mila (Quebec AI Institute) for providing computational resources and support that contributed to these research results.

\small
\bibliographystyle{plain}
\bibliography{neurips_2021.bib}

\begin{thebibliography}{10}

\bibitem{Bustos2020PadChestAL}
A.~Bustos, A.~Pertusa, J.~M. Salinas, and M.~Iglesia-Vay{\'a}.
\newblock Padchest: A large chest x-ray image dataset with multi-label
  annotated reports.
\newblock {\em Medical image analysis}, 66:101797, 2020.

\bibitem{Cohen2020OnTL}
Joseph~Paul Cohen, Mohammad Hashir, Rupert Brooks, and Hadrien Bertrand.
\newblock {On the limits of cross-domain generalization in automated X-ray
  prediction}.
\newblock In {\em Medical Imaging with Deep Learning}, 2020.

\bibitem{Cohen2020xrv}
Joseph~Paul Cohen, Joseph Viviano, Paul Morrison, Rupert Brooks, Mohammad
  Hashir, and Hadrien Bertrand.
\newblock {TorchXRayVision: A library of chest X-ray datasets and models}.
\newblock {\em https://github.com/mlmed/torchxrayvision}, 2020.

\bibitem{Goodfellow2015DeepL}
I.~Goodfellow, Yoshua Bengio, and Aaron~C. Courville.
\newblock Deep learning.
\newblock {\em Nature}, 521:436--444, 2015.

\bibitem{Hao2016DeepL}
Xing Hao, Guigang Zhang, and Shang Ma.
\newblock Deep learning.
\newblock {\em Int. J. Semantic Comput.}, 10:417--, 2016.

\bibitem{Irvin2019CheXpertAL}
Jeremy~A. Irvin, Pranav Rajpurkar, M.~Ko, Yifan Yu, Silviana Ciurea-Ilcus,
  Chris Chute, H.~Marklund, Behzad Haghgoo, Robyn~L. Ball, K.~Shpanskaya,
  J.~Seekins, D.~Mong, S.~Halabi, J.~Sandberg, Ricky~H Jones, D.~Larson,
  C.~Langlotz, B.~Patel, M.~Lungren, and A.~Ng.
\newblock Chexpert: A large chest radiograph dataset with uncertainty labels
  and expert comparison.
\newblock In {\em AAAI}, 2019.

\bibitem{Johnson2019MIMICCXRAL}
Alistair E.~W. Johnson, T.~Pollard, Seth~J. Berkowitz, Nathaniel~R. Greenbaum,
  M.~Lungren, Chih ying Deng, R.~Mark, and S.~Horng.
\newblock Mimic-cxr: A large publicly available database of labeled chest
  radiographs.
\newblock {\em ArXiv}, abs/1901.07042, 2019.

\bibitem{Kingma2015AdamAM}
Diederik~P. Kingma and Jimmy Ba.
\newblock Adam: A method for stochastic optimization.
\newblock {\em CoRR}, abs/1412.6980, 2015.

\bibitem{MorenoTorres2012AUV}
J.~G. Moreno-Torres, T.~Raeder, R.~Ala{\'i}z-Rodr{\'i}guez, N.~Chawla, and
  F.~Herrera.
\newblock A unifying view on dataset shift in classification.
\newblock {\em Pattern Recognit.}, 45:521--530, 2012.

\bibitem{Pan2010ASO}
Sinno~Jialin Pan and Qiang Yang.
\newblock A survey on transfer learning.
\newblock {\em IEEE Transactions on Knowledge and Data Engineering},
  22:1345--1359, 2010.

\bibitem{Prechelt2012EarlyS}
L.~Prechelt.
\newblock Early stopping - but when?
\newblock In {\em Neural Networks: Tricks of the Trade}, 2012.

\bibitem{QuioneroCandela2009DatasetSI}
Joaquin Quionero-Candela, Masashi Sugiyama, Anton Schwaighofer, and Neil
  Lawrence.
\newblock Dataset shift in machine learning.
\newblock 2009.

\bibitem{Russakovsky2015ImageNetLS}
Olga Russakovsky, J.~Deng, Hao Su, J.~Krause, S.~Satheesh, S.~Ma, Zhiheng
  Huang, A.~Karpathy, A.~Khosla, Michael~S. Bernstein, A.~Berg, and Li~Fei-Fei.
\newblock Imagenet large scale visual recognition challenge.
\newblock {\em International Journal of Computer Vision}, 115:211--252, 2015.

\bibitem{Wang2017ChestXRay8HC}
Xiaosong Wang, Yifan Peng, Le~Lu, Zhiyong Lu, M.~Bagheri, and R.~Summers.
\newblock Chestx-ray8: Hospital-scale chest x-ray database and benchmarks on
  weakly-supervised classification and localization of common thorax diseases.
\newblock {\em 2017 IEEE Conference on Computer Vision and Pattern Recognition
  (CVPR)}, pages 3462--3471, 2017.

\bibitem{Wong2016UnderstandingDA}
Sebastien~C. Wong, A.~Gatt, V.~Stamatescu, and M.~D. McDonnell.
\newblock Understanding data augmentation for classification: When to warp?
\newblock {\em 2016 International Conference on Digital Image Computing:
  Techniques and Applications (DICTA)}, pages 1--6, 2016.

\bibitem{Yosinski2014HowTA}
J.~Yosinski, J.~Clune, Yoshua Bengio, and Hod Lipson.
\newblock How transferable are features in deep neural networks?
\newblock {\em ArXiv}, abs/1411.1792, 2014.

\end{thebibliography}


\end{document}